\documentclass[crcready]{iosart2x}

\usepackage{subfiles}
\usepackage{supertabular}
\usepackage{soul}
\usepackage{fancyvrb}
\graphicspath{ {./images/} }
\usepackage{array}
\newcolumntype{H}{>{\setbox0=\hbox\bgroup}c<{\egroup}@{}}

\firstpage{1}
\lastpage{22}
\volume{1}
\pubyear{0000}

\begin{document}
\makeatletter
\let\put@numberlines@box\relax
\makeatother

\begin{frontmatter}

\title{ExtruOnt: An ontology for describing a type of manufacturing machine for Industry 4.0 systems}
\runningtitle{ExtruOnt: An Ontology for describing a type of manufacturing machine for Industry 4.0 systems}


\author[A]{\inits{N.}\fnms{V{\'i}ctor~Julio} \snm{Ram{\'i}rez-Dur{\'a}n}\ead[label=e1]{victorjulio.ramirez@ehu.eus}%
\thanks{Corresponding author. \printead{e1}.}},
\author[A]{\inits{N.N.}\fnms{Idoia} \snm{Berges}\ead[label=e2]{idoia.berges@ehu.eus}}
and
\author[A]{\inits{N.-N.}\fnms{Arantza} \snm{Illarramendi}\ead[label=e3]{a.illarramendi@ehu.eus}}
\runningauthor{V. Ram{\'i}rez et al.}
\address[A]{Department of Languages and Information Systems, \orgname{University of the Basque Country UPV/EHU}, Manuel Lardizabal Ibilbidea, 1, 20018 Donostia-San Sebasti{\'a}n, \cny{Spain}\printead[presep={\\}]{e1,e2,e3}}

\begin{review}{editor}
\reviewer{\fnms{First} \snm{Editor}\address{\orgname{University or Company name}, \cny{Country}}}
\reviewer{\fnms{Second} \snm{Editor}\address{\orgname{University or Company name}, \cny{Country}}}
\end{review}
\begin{review}{solicited}
\reviewer{\fnms{First Solicited} \snm{Reviewer}\address{\orgname{University or Company name}, \cny{Country}}}
\reviewer{\fnms{Second Solicited} \snm{Reviewer}\address{\orgname{University or Company name}, \cny{Country}}}
\end{review}
\begin{review}{open}
\reviewer{\fnms{First Open} \snm{Reviewer}\address{\orgname{University or Company name}, \cny{Country}}}
\reviewer{\fnms{Second Open} \snm{Reviewer}\address{\orgname{University or Company name}, \cny{Country}}}
\end{review}

\begin{abstract}
Semantically rich descriptions of manufacturing machines, offered in a machine-interpretable code, can provide interesting benefits in Industry 4.0 scenarios. However, the lack of that type of descriptions is evident. In this paper we present the development effort made to build an ontology, called \textit{ExtruOnt}, for describing a type of manufacturing machine, more precisely, a type that performs an extrusion process (extruder). Although the scope of the ontology is restricted to a concrete domain, it could be used as a model for the development of other ontologies for describing manufacturing machines in Industry 4.0 scenarios. 

The terms of the \textit{ExtruOnt} ontology provide different types of information related with an extruder, which are reflected in distinct modules that constitute the ontology. Thus, it contains classes and properties for expressing descriptions about 
components of an extruder, 
spatial connections, features, and 3D representations of those components, and finally the sensors used to capture indicators about the performance of this type of machine.
The ontology development process has been carried out in close collaboration with domain experts.

\end{abstract}

\begin{keyword}
\kwd{Ontology}
\kwd{Extruder}
\kwd{Industry 4.0}
\kwd{Smart Manufacturing}
\end{keyword}

\end{frontmatter}


\section{Introduction}
Different initiatives and strategies are emerging in the 4th Industrial revolution (Industry 4.0) that is currently being experienced in the manufacturing sector. Mainly they address, on the one hand, the compilation of manufacturing records of products, with data about their history, state, quality and characteristics, and on the other hand,  the application of manufacturing intelligence to those records, so that the exploitation of those data allows manufacturers to predict, plan and manage specific circumstances in order to optimize their production. Those initiatives enable important business opportunities for the manufacturers.

Moreover, the appropriate design and implementation of such initiatives requires an innovation effort by deploying, among others, mechatronics for advanced manufacturing systems, manufacturing strategies, knowledge-workers and modelling, simulation and forecasting methods and tools \cite{effra}. Concerning modeling, a lack of sound descriptions of manufacturing machines that happen to be accessible, interoperable, and reusable can be identified nowadays.
Thus, in order to alleviate that existing shortage we have developed an ontology for providing detailed descriptions of a real manufacturing machine type (called extruder) that performs an extrusion process\footnote{In which some material is forced through a series of dies in order to create a desired shape.}. 
We have not found any other ontology concerning extruders, however, we believe that the ontology-based description of different manufacturing machine types 
can contribute significantly to the development of the Industry 4.0.

The purpose of this paper is to present the \textit{ExtruOnt} ontology. It includes terms to describe 1) the \textit{main components} of an extruder (e.g. the drive system), 2) the \textit{spatial connections} between the extruder components (e.g. the filter is externally connected to the barrel), 3) the \textit{different features} of the components (e.g. the power consumption of the motor is 40.5 kWh), 4) the \textit{3D description} of the position of the components (e.g. the feed hopper is located at point \textit{q(0,0,-1)} in a 3D canvas), and, 5) the \textit{sensors} that need to be used to capture indicators about the performance of that extruder (e.g the temperature sensor that captures the melting temperature of the polymer).

The \textit{ExtruOnt} ontology has been implemented using OWL 2\footnote{\url{https://www.w3.org/TR/owl2-overview/}} and  the Prot\'eg\'e\footnote{\url{https://protege.stanford.edu/}} \cite{noy2003protege} development environment. 
\textit{ExtruOnt} is in line with concepts included in  an ontology-based context model for industry presented in \cite{GIUSTOZZI2018675} and is aligned with several ontologies: the DUL ontology \footnote{\url{http://ontologydesignpatterns.org/wiki/Ontology:DOLCE+DnS\_Ultralite}}, which models physical contexts; 
the MASON ontology, an upper ontology for representing the core concepts of the manufacturing domain \cite{Lemaignan06};
SAREF4INMA \cite{saref4inma}, a  SAREF \cite{Etsi17} extension for semantic interoperability in the industry and manufacturing domain; the GeoSPARQL ontology,  which incorporates descriptions about Region Connection Calculus (RCC) \cite{battle2011geosparql}; the OM\footnote{https://enterpriseintegrationlab.github.io/icity/OM/doc/index-en.html} ontology, the largest unit ontology \cite{rijgersberg2013ontology}; the 3D Modeling Ontology (3DMO), which maps the entire XSD-based vocabulary of the industry standard X3D\footnote{http://www.web3d.org/what-x3d-graphics} (ISO/IEC 19775-19777) to OWL 2 \cite{sikos2017novel} and with the SOSA/SSN,  which defines general concepts about sensors \cite{haller2018sosa}.

Apart from the interest that the \textit{ExtruOnt} ontology has in itself, the main contributions of the \textit{ExtruOnt} ontology are the following: 1) Reusability.  Its modular design facilitates the task of developing other ontologies for different types of manufacturing machines. The module that describes the components of an extruder could be replaced by another module that would describe another type of manufacturing machine, while alignments with other modules should be adapted to meet the requirements of the new type of machine. Moreover, the defined alignments of \textit{ExtruOnt} ontology with upper ontologies such as DUL and MASON facilitate the task of modeling different manufacturing operations (e.g. customer orders, production plans); 2) Expressiveness of Spatial Connections. It incorporates a hierarchical description  of  possible relations in Region Connection Calculus and some custom-defined ones. Dealing with all those descriptions, more specific spatial relations can be defined and thus fine-grained results for questions can be provided.

Finally, the use of the \textit{ExtruOnt} ontology as the core element of ontology-based systems, developed for Smart Manufacturing scenarios, can bring several benefits. For example, the development of an ontology-based Visual Query System will bring the following benefits to the different types of workers of a manufacturing plant:

\begin{itemize}
\item \textit{Novice workers.}  The 3D rendering of an extruder machine
obtained from descriptions in the ontology
will allow novice workers to familiarize themselves with the extrusion process due to its similarity to reality.
\item \textit{Product Designers.}  The descriptions referring to the components of the extruder as well as the constraints regarding their spatial connections, positioning and features contained in the ontology will facilitate product designers the task of creating customized 3D images of extruder machines. 
\item \textit{Domain experts.} Ontology-based annotation of data captured by sensors will allow domain experts 
to perform an assisted exploration of data.
\end{itemize}

In the rest of this paper, we present first, distinct approaches that have been defined in the literature, related to two aspects considered during the development process of \textit{ExtruOnt}: existing ontologies and ontology evaluation techniques. Then, we show some methodologies that have been proposed to adequately develop ontologies. Next, we illustrate the steps that we followed to develop the \textit{ExtruOnt} ontology using the NeOn methodology \cite{Neon2012} and the modules that constitute \textit{ExtruOnt}. Later, we show the results of the evaluation process carried out considering two goals: domain coverage and quality of modeling. We finish with some conclusions and future work.

\section{Related work} \label{relatedwork}
 In the specialized literature several ontologies related to the Smart Manufacturing area can be found. Those ontologies were defined with distinct purposes and, therefore, describe different types of information related to that area. For example, the PSL (Process Specification Language) ontology \cite{Gruninger09} includes fundamental concepts for representing manufacturing processes. The foundational elements of the core of the PSL ontology are four primitive classes (\textit{activity}, \textit{activity-occurrence}, \textit{timepoint}, \textit{object}), three primitive relations (\textit{participates-in}, \textit{before}, \textit{occurrence-of}) and two primitive functions (\textit{beginof}, \textit{endof}). The MASON (Manufacturing's Semantics  Ontology) ontology \cite{Lemaignan06} is an upper ontology for representing what authors consider the core concepts of the manufacturing domain: products, processes and resources. As a result, the main classes of MASON are \textit{Entity} (for specifying the product), \textit{Operation} (for describing all processes linked to manufacturing) and \textit{Resource} (for representing concepts regarding machine-tools, tools, human resources and geographic resources). The SIMPM (Semantically Integrated Manufacturing Planning Model) ontology \cite{Sormaz2019} is an upper ontology that models the fundamental constraints of manufacturing process planning: manufacturing activities and resources, time and aggregation. MaRCO (Manufacturing Resource Capability Ontology) \cite{Jarvenpaa2019} defines capabilities of manufacturing resources. Its main class is \textit{Capability}, which is specialized to cover both, simple capabilities (e.g. \textit{Fixturing}, \textit{SpinningTool}) and combined capabilities (those that require a combination of two or more simple capabilities, e.g. \textit{PickAndPlace}, which requires \textit{FingerGrasping} or \textit{Vacuum Grasping}, \textit{Moving} and \textit{Releasing}). The MSDL (Manufacturing Service Description Language) ontology \cite{ameri2006upper} allows to describe manufacturing services. More precisely, a \textit{Manufacturing Service} is seen as a \textit{Service} that is provided by a \textit{Supplier} and that has some \textit{Manufacturing Capability}, which is enabled by some \textit{Manufacturing Resource} and delivered by some \textit{Manufacturing Process}.
The P-PSO (Politecnico di MilanoProduction Systems) ontology \cite{Garetti12} considers three aspects in the manufacturing domain: the physical aspect (the material definition of the system), the technological aspect (the operational view of the system) and the control aspect (the management activities), for information exchange, design, control, simulation and other applications. Thus, its main classes are \textit{component}, \textit{operation} and \textit{controller}, which model the aforementioned three aspects, as well as \textit{part}, \textit{operator} and \textit{subsystem}.
OntoSTEP (ONTOlogy of Standard for the Exchange of Product model data) \cite{OntoStep12} allows the description of product information mainly related to geometry.
MCCO (Manufacturing Core Concepts Ontology) \cite{MCCO} focuses on interoperability across the production and design domains of product lifecycle. It provides some core classes in categories such as \textit{ManufacturingProcess}, \textit{ManufacturingFacility}, \textit{ManufacturingResource} and \textit{Feature}.
Finally, SAREF4INMA \cite{Etsi17} pursues favouring interoperability with industry standards. Some of its main classes are \textit{ProductionEquipment}, \textit{Factory}, \textit{Item} and \textit{MaterialCategory}.

 Although some of the mentioned ontologies contain some general terms for representing the concept of industrial machine (e.g. \textit{Machine-tool} in MASON, \textit{Device} in MarCO, \textit{ProductionEquipment} in SAREF4INMA), further specialization and characterization are needed for fitting our goal, that is, for describing specific industrial machine types with a fine-grained detail, and more particularly, extruder machines. The search on different ontology repositories (e.g. LOV \cite{LOV2017}, Swoogle \cite{swoogle2004}, ODP \cite{gangemi2010ontology}) for an ontology that covered this domain yielded unsuccessful, and for that reason we built the \textit{ExtruOnt} ontology following a well-established methodology. 

Furthermore, considering the relevance of evaluating the quality and correctness of an ontology once it has been built, several evaluation approaches have been proposed in the specialized literature depending on the evaluation goal.
The NeOn guidelines for carrying out the ontology evaluation activity \cite{Neonevaluation} identify the following goals of evaluation: \textit{domain coverage}, \textit{quality of modeling}, \textit{suitability for an application/task} and \textit{adoption and use}. Then, specific evaluation approaches need to be chosen depending on the selected goals. These approaches include, among others, comparing to a gold standard ontology \cite{Maedche02}, comparing to unstructured or informal data \cite{Brewster04}, using human assessments \cite{Tello04}, and using reasoners to assess the logical correctness of the ontology \cite{Horridge09}. Another relevant work in the area of ontology evaluation is the one in \cite{Vrandecic2009}, where a common framework that considers quality criteria for aspects of ontology evaluation is presented. More precisely, it identifies the following criteria: \textit{accuracy},  \textit{adaptability}, \textit{clarity}, \textit{completeness}, \textit{computational efficiency}, \textit{conciseness}, \textit{consistency} and  \textit{organizational fitness}. In the case of the proposed \textit{ExtruOnt} ontology, some aspects considered in those works were taken into account during the evaluation process (see section \ref{eva}).

\section{Design Methodologies} \label{designmethodology}
Different methodologies such as On-To-Knowledge \cite{Sure2004}, Diligent \cite{Pinto2004} and NeOn \cite{Neon2012} can be found in the literature to adequately develop well-founded ontologies. On-To-Knowledge proposes a knowledge meta process consisting of five steps: \textit{feasibility study} to determine whether to begin the actual development of the ontology; \textit{kickoff}, where the requirements are specified and a semi-formal ontology description is developed; \textit{refinement}, where the target ontology is obtained by refining and formalizing the semi-formal one;  \textit{evaluation}, where the evaluation of the ontology is done; and  \textit{application and evolution}, where the ontology is applied in the target system and maintained. On-To-Knowledge suggests reusing ontologies in the kickoff step if available, but does not provide any guidelines for it. Moreover, it does not deal with non-ontological resources nor other ontological resources such as ontology design patterns. Diligent proposes a process for a distributed development of ontologies that comprises five main steps: \textit{build}, where an initial version of the ontology is built by different stakeholders such as domain experts, users, and knowledge and ontology engineers; \textit{local adaptation}, where users adapt the ontology for their own purposes; \textit{analysis}, where a control board analyses the local versions to detect similarities and decide which changes and requests are added to the next shared version of the ontology; \textit{revision}, where the board revises the new version of the shared ontology; and \textit{local update}, where users can update their local ontologies with information from the new version. This methodology does not detail the series of activities that should be followed during the \textit{build} step, and moreover, it does not include guidelines for using neither ontological nor non-ontological resources in the development process. The NeOn methodology describes a set of nine scenarios that may occur when building an ontology, along with a list  of  activities  that  should  be  carried out in each scenario. Tightly related to those scenarios, it presents two ontology network life cycle models (waterfall and iterative-incremental) with several versions. The basic version is the Four-phase model, which includes the following phases: \textit{initiation}, where the requirements are specified; \textit{design}, where both an informal and a formal model of the ontology are created;  \textit{implementation}, where the formal model is implemented in an ontology language; and \textit{maintenance}, where the ontology is used until errors or missing knowledge are detected. The NeOn methodology places special emphasis on reusing and re-engineering both ontological and non-ontological knowledge resources. Thus, more detailed versions of the basic model (e.g Five-phase model, Six-phase + Merging model) include as well one or more of the following phases, resulting in a variety of paths to develop an ontology: \textit{reuse}, where existing ontological or non-ontological resources are added to the model; \textit{re-engineering}, where those resources are modified to serve to the intended purpose; and \textit{merging}, where ontologies are merged or alignments are established among ontological resources. The methodology includes thorough guidelines on how to perform all the mentioned activities.

\section{Development of the \textit{ExtruOnt} ontology} \label{method}
\begin{figure}
    \centering
    \includegraphics[width=0.97\columnwidth]{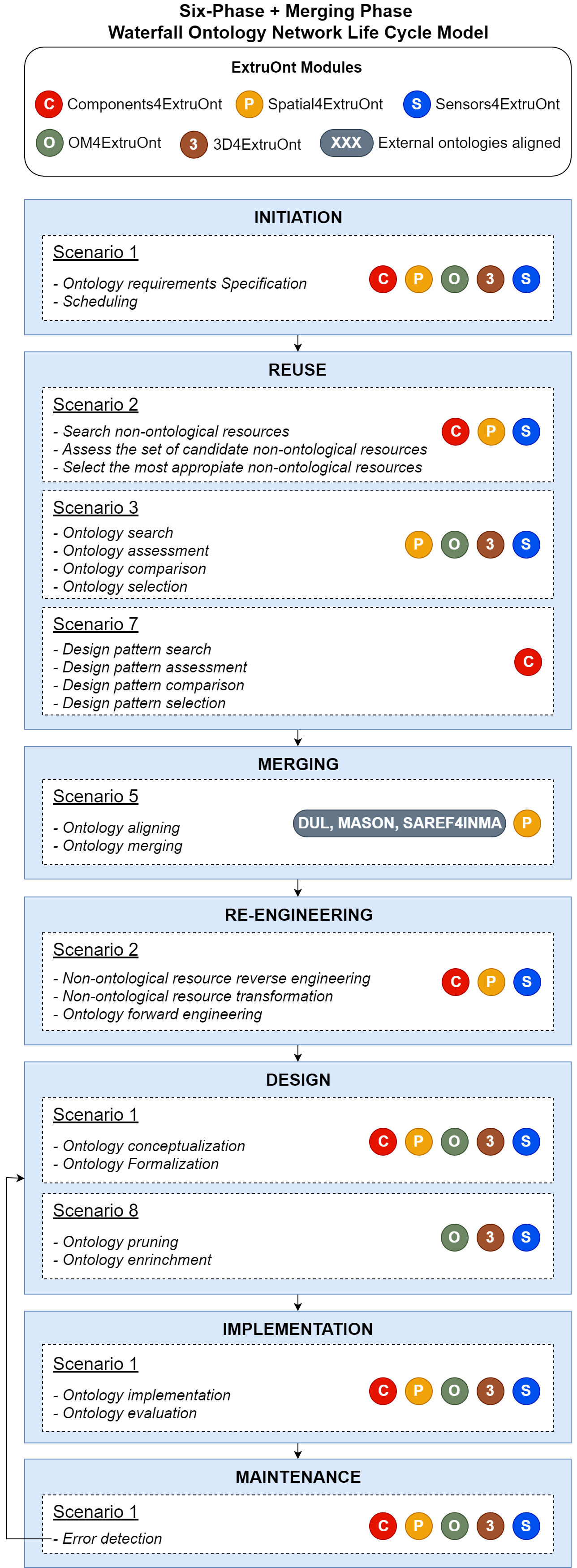}
    \caption{The Six-Phase + Merging Phase Waterfall Ontology Network Life Cycle Model along with scenarios, activities and \textit{ExtruOnt} modules.}
    \label{fig:NeOnModelScenarioActivity}
\end{figure}

In order to develop the \textit{ExtruOnt} ontology we selected the NeOn methodology. In our opinion, NeOn beats the other methodologies in these two aspects: on the one hand, the variety of scenarios that it takes into account, which results in a more flexible methodology, and on the other hand, the great detail in the description of the activities that need to be carried out when building the ontology. Furthermore, due to the requirements of \textit{ExtruOnt}, which include reuse of ontological and no-ontological resources, re-engineering, merging, aligning with domain ontologies, implementation and evaluation among others, its development process fits with the Six-Phase + Merging Phase Waterfall Ontology Network Life Cycle Model.
In figure \ref{fig:NeOnModelScenarioActivity} the phases of the aforementioned life cycle model along with scenarios, activities and modules of the \textit{ExtruOnt} ontology involved in each scenario are indicated. These modules and their purpose are explained in section \ref{mod}. The different phases of the life cycle model are explained below.

\subsection{Initiation}

In collaboration with the R\&D director of a company that manufactures extruder machines,
we created the Ontology Requirements Specification Document (ORSD) that contains among others, the purpose of the \textit{ExtruOnt} ontology, its scope and the Competency Questions (CQs), see Table~\ref{t1}.
After a detailed analysis of those questions, it was noticed that they referred to five different dimensions regarding information related to extruders. Thus, the questions were classified in the following five groups, one for each dimension: the components of an extruder, the spatial connections between those components, their features, their 3D description and the sensors that capture information about several indicators (Scenario 1).

\begin{table*}[htb]
    \caption{Summary of the Ontology Requirements Specification Document for \textit{ExtruOnt}} \label{t1}
    \begin{tabular}{l p{15cm}}
        \hline
        1. &\textbf{Purpose} \\
        & The purpose of the \textit{ExtruOnt} ontology is to provide a reference model for the physical representation of extruder machines and the time series data gathered from their sensors, allowing to describe the extruder components, their position with respect to other components and the data obtained from sensing devices.\\[4pt]
        2. &\textbf{Scope} \\
        & The ontology will focus on general purpose extruder machines.\\[4pt]
        3. &\textbf{Implementation language} \\
        & The ontology has to be implemented in a formalism that allows classification of classes and realization between instances and classes.\\[4pt]
        4. &\textbf{Intended users} \\
        & \vspace{-4mm}\begin{itemize}
            \item \emph{User 1:} Novice workers.
            \item \emph{User 2:} Product designers.
            \item \emph{User 3:} Domain Experts.
          \end{itemize}\vspace{-5mm}
        \\[4pt]
        5. &\textbf{Intended uses} \\
        & \vspace{-4mm}\begin{itemize}
            \item \emph{Use 1:} To describe different models of extruders.
            \item \emph{Use 2:}  To help the process of identifying the extruder components and their location.
            \item \emph{Use 3:} To help to select the optimal extruder for a specific product.
            \item \emph{Use 4:} To recognize differences between extruder models.
            \item \emph{Use 5:} To improve user interaction with the different sensing devices in the extruder and the gathered data.
          \end{itemize}\vspace{-5mm}
        \\[5pt]
        6. &\textbf{Ontology requirements} \\
        & (6.a) Non-functional requirements (not applicable) \\
        & (6.b)  Functional requirements: Groups of competency questions \\
        & \vspace{-4mm}\begin{itemize}
            \item \emph{CQG1:} Extruder components-related competency questions:
                \begin{itemize}
                    \item CQ1.1: How many heater bands does the extruder E01 have?
                    \item CQ1.2: What kind of extrusion head does the extruder E02 have?
                    \item CQ1.3: Is the machine E03 a single or double screw extruder?
                    \item CQ1.4: Is the extruder E04 powered by an AC motor?
                    \item CQ1.5: Is this extruder E05 suitable to process plastic pellets?
                    \item CQ1.6: Can the extruder E06 process multiple polymers?
                    \item \dots
                \end{itemize}
            \item \emph{CQG2:} Spatial connections-related competency questions:
                \begin{itemize}
                    \item CQ2.1: With which components are the filters FIL01 connected?
                    \item CQ2.2: Which components overlap the barrel BAR01?
                    \item CQ2.3: Which components are disconnected from the motor M01?
                    \item CQ2.4: Which components are monitored in the drive system DS01?
                    \item CQ2.5: How many sensors does the barrel BAR02 have?
                \item \dots
                \end{itemize}
            \item \emph{CQG3:} Features-related competency questions:
                \begin{itemize}
                    \item CQ3.1: What is the diameter of the barrel BAR03?
                    \item CQ3.2: What are the optimal operating conditions of the screw SCR01?
                    \item CQ3.3: What is the maximum torque produced by the motor M02?
                    \item CQ3.4: Does the extruder E07 fit in a space 3 meters wide by 5 meters long?
                    \item CQ3.5: What is the bottles-per-hour production rate of the extruder E08?
                \item \dots
                \end{itemize}
            \item \emph{CQG4:} 3D positioning-related competency questions:
                \begin{itemize}
                    \item CQ4.1: Which components of extruder E11 can not be located in a 3D canvas?
                    \item CQ4.2: What are the modeling and position of the feed hopper FH01?
                    \item \dots
                \end{itemize}
          \end{itemize}\vspace{-5mm}
        \\[3pt]
        \hline
    \end{tabular}
\end{table*}

\addtocounter{table}{-1}

\begin{table*}[htb]
    \caption{Continued}
    \begin{tabular}{l p{15cm}}
        \hline
        & \vspace{-4mm}\begin{itemize}
            \item \emph{CQG5:} Sensors and observations-related competency questions:
                \begin{itemize}
                    \item CQ5.1: What properties are observed by the sensors located in the extrusion head EH01?
                    \item CQ5.2: What is the unit of measurement used by the motor consumption sensor MCS01?
                    \item CQ5.3: Where is the melting temperature sensor located in extruder E08?
                    \item CQ5.4: What is the identifier of the temperature sensor in extrusion head EH02?
                    \item CQ5.5: When was the first and last observation made by sensor SN01?
                    \item CQ5.6: What was the average, maximum and minimum value of the observations in a day for the sensor SN02?
                    \item CQ5.7: How many observations from torque sensor SN03 are outside the optimal values?
                    \item CQ5.8: how long was the maximum period of extruder E09 inactivity during the last week?
                    \item CQ5.9: At what times during August 21st, 2018 and August 22nd, 2018 did the melting temperature exceed the maximum optimal operational value in extruder E10?
                \item \dots
                \end{itemize}
          \end{itemize}\vspace{-5mm}
        \\[6pt]
        7. &\textbf{Pre-glossary of terms} \\
        & Extruder, feed system, observation, sensor, tangential proper part, measure, 3D canvas ...\\[3pt]
        \hline
    \end{tabular}
\end{table*}

\subsection{Reuse}

Due to the fact that the search for an ontology that covered all these dimensions was unsuccessful, we focused on searching both ontological and non-ontological resources for each dimension.

In this subsection, we present the non-ontological and ontological resources used to describe the aforementioned dimensions. 

\begin{itemize}
    \item \textit{Components of an extruder}: 
    In order to describe the components, we relied on the one hand, on non-ontological resources existing in the specialized literature and mainly in a full chapter dedicated to the extruder and its equipment that appears in \cite{giles2004extrusion}. Moreover, due to the complexity of the extrusion head, another non-ontological resource was used as a reference to represent the features of this component. In \cite{sikora2008design}, a thorough explanation of the extrusion head design and applications is presented, categorizing the extrusion head depending on the position and the type of extrudate obtained (Scenario 2). On the other hand, the PartOf\footnote{\url{http://ontologydesignpatterns.org}} ontology design pattern was selected in order to specify parthood between the extruder and its components, as well as between different parts that constitute each component (Scenario 7).
    
    \item \textit{Spatial connections between components}: In the specialized literature can be found the Region Connection Calculus (RCC) \cite{randell1992spatial,Cohn1997}, which is intended to represent the spatial relations between objects and facilitate reasoning over those relations. There are multiple representations of the RCC. The main one is RCC8, which consists of 8 basic relations that are possible between two regions. Different ontologies have tried to represent the RCC descriptions (GeoSPARQL\cite{battle2011geosparql}, Spatial Relations Ontology\footnote{\url{http://data.ordnancesurvey.co.uk/ontology/spatialrelations/}}, NeoGeo Spatial Ontology \footnote{\url{http://geovocab.org/doc/neogeo/}}). 
    We selected the GeoSPARQL ontology, which models the RCC8 relations, because it is the base for the other spatial ontologies (Scenario 3).
    
    \item \textit{Features of the components}:
    Based on a work that evaluates ontologies of measurements \cite{keil2018comparison}, two ontologies were considered:
    QUDT\footnote{http://www.linkedmodel.org/catalog/qudt/1.1/index.html} \cite{hodgson2011qudt} and OM\footnote{https://enterpriseintegrationlab.github.io/icity/OM/doc/index-en.html} \cite{rijgersberg2013ontology}. QUDT is the result of a NASA-sponsored initiative to formalize Quantities, Units of Measure, Dimensions and Types, and it is categorized as a medium sized ontology. OM is an ontology that allows to
    model concepts and relations in the context of food research and it was the largest unit ontology compared. In the aforementioned evaluation, multiple issues were found in QUDT ontology like reasoning impossibility, duplicated units, wrong specifications, typing errors, etc. Moreover, only English labels were added and, according to the article, the reported issues remain unsolved. On the other hand, OM shared some issues with QUDT like reasoning impossibility, wrong dimension values, typing errors, but the reported issues have been corrected and labelling can be found in Dutch and Chinese for a subset of individuals. Equally important, more concepts can be found in OM, so this was the selected ontology (Scenario 3).
    
    \item \textit{3D representation of components}: We selected the 3D Modeling Ontology (3DMO) \cite{sikos2017novel} because this ontology maps the entire XSD-based vocabulary of the industry standard X3D\footnote{http://www.web3d.org/what-x3d-graphics} (ISO/IEC 19775-19777) to OWL 2. Therefore, it can be used for the representation, annotation, and efficient indexing of 3D models (Scenario 3). 
    
    \item \textit{Sensors for capturing information about indicators}: We did not find any ontological resource that defines the specific types of sensors that are used to monitor extruders. However, the well known SOSA/SSN\cite{haller2018sosa} ontology defines general concepts about sensors, which can be specialized with information obtained from non-ontological resources about extruders \cite{giles2004extrusion} to reflect the specificities of the extrusion domain (scenario 3).
\end{itemize}

\subsection{Merging}
To guarantee semantic interoperability, the \textit{ExtruOnt} ontology is aligned with other domain ontologies such as: 1) DUL, an upper ontology created to provide a set of concepts to facilitate interoperability among ontologies; 2) MASON, an upper ontology for representing the core concepts of the manufacturing domain
and 3) SAREF4INMA, a SAREF extension for industry and manufacturing (scenario 5). The selection of these ontologies was carried out taking into account different key factors such as domain, use, maintenance, acceptance, popularity and coverage. For example, in the selection of MASON, other different ontologies were considered: MaRCO, whose approach is oriented to machine capabilities and, thus, out of our scope; MSDL, with a large amount of concepts focused on processes and resources but leaving products aside; SIMPM, with few concepts and focused only on the processes; and finally, PSL, P-PSO, MCCO and OntoSTEP whose OWL definitions could not be found. On the contrary, MASON defines a meaningful categorization of products, processes and resources, it has been widely reviewed \cite{cao2018ontologies} and it is currently available. 
The terms used in the ontology alignment are presented in section \ref{c4e}.

Concerning to the spatial connection between components, we realized that using only the GeoSPARQL ontology was not sufficient for answering competency question CQ2.2. Thus, a twofold approach was used: in addition to the  GeoSPARQL ontology, information about other RCC spatial relations obtained from the aforementioned non-ontological RCC resources was incorporated (scenario 5). 
    
\subsection{Re-engineering}
A re-engineering process was carried out to transform the non-ontological resources mentioned previously into conceptual models, analyzing the structure of the resource (chapters, subsections, connections, order, etc.). Once the conceptual model for each resource had been created, they were used as input of the design phase. (Scenario 2).

\subsection{Design}
 The modularization of ontologies facilitates the development, reuse and maintenance of an ontology. In addition, it conforms to the dimensionality approach obtained from the ORSD analysis. Therefore, each of the five dimensions was represented through a module: the components of an extruder (\textit{components4ExtruOnt}), the spatial connections between those components (\textit{spatial4ExtruOnt}), their features (\textit{OM4ExtruOnt}), their 3D description (\textit{3D4Extru-}\textit{Ont}) and the sensors that capture information about several indicators (\textit{sensors4ExtruOnt}), which altogether form the \textit{ExtruOnt}\footnote{\url{http://bdi.si.ehu.es/bdi/ontologies/ExtruOnt/ExtruOnt.owl}} ontology (Scenario 1). The key features of each module are presented in depth in section \ref{mod}. 
 
 OM, SOSA/SSN and 3DMO ontologies contain a wide range of concepts that belong to the domains they represent, however, due to the specific domain we wanted to model, a pruning process was carried out for these ontologies keeping only those concepts and descriptions that are relevant, favoring lightness, cleanliness and maintenance of the ontology  (Scenario 8). Additionally, the pruned SOSA/SSN ontology was enriched with specialized concepts drawn from the conceptual model (see section \ref{se4e}).

\subsection{Implementation}
A formal model expressed in a Description Logic was generated and implemented in OWL 2 DL Web Ontology Language using Prot{\'e}g{\'e} \cite{noy2003protege} (Scenario 1). Later, a wide evaluation of the ontology was done which is presented in section \ref{eva}, describing the different considered approaches.

\subsection{Maintenance}
The maintenance phase is currently undergoing. Once an error is detected, the ontology will be taken to the design phase to be corrected, as stipulated in the Waterfall ontology network life cycle model.

\section{Ontology modules} \label{mod}
\begin{figure*}[h]
    \centering
    \includegraphics[width=0.8\textwidth]{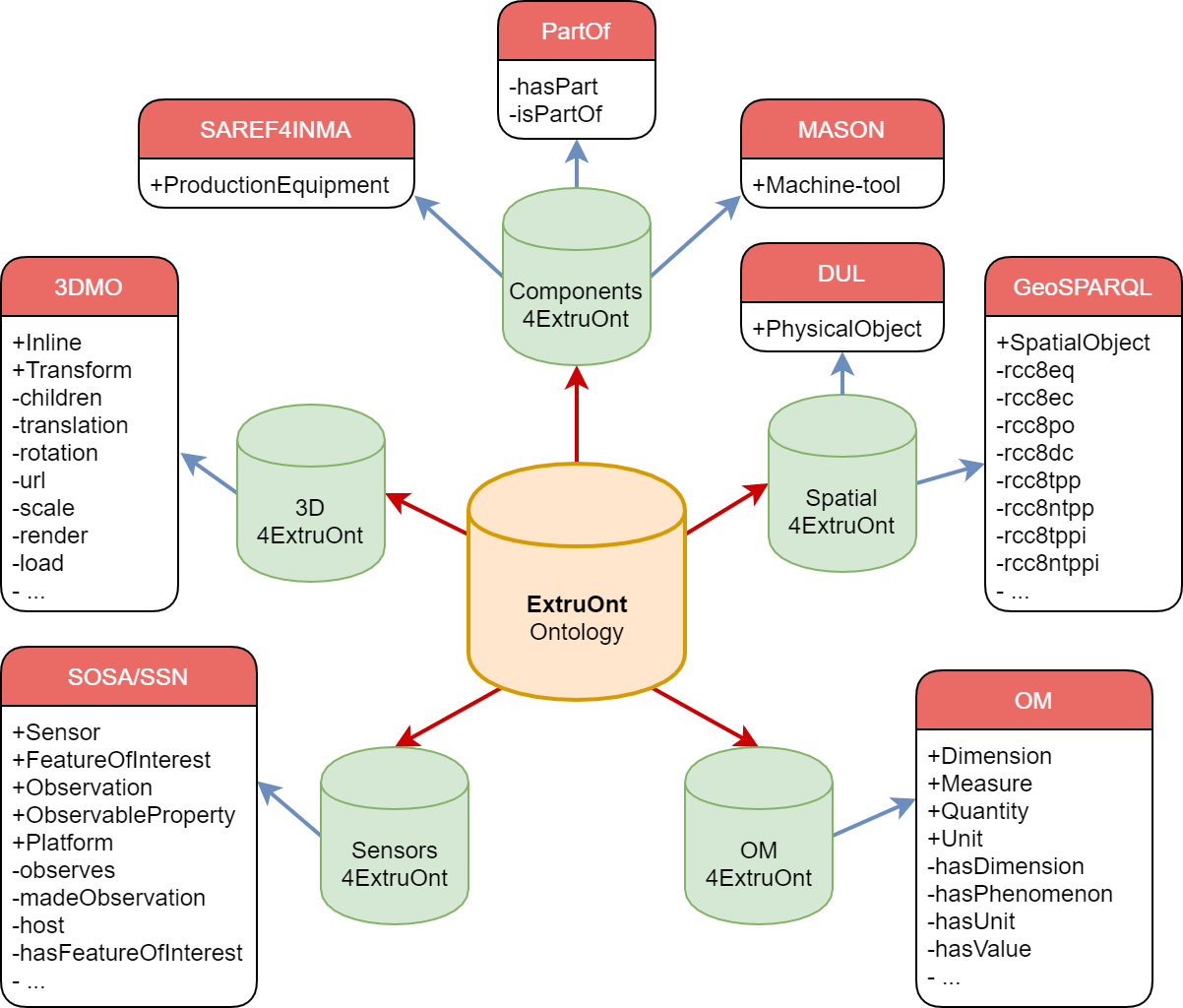}
    \caption{\textit{ExtruOnt} ontology diagram showing the reuse of terms from other domain ontologies.}
    \label{fig:ExtruOntDiag}
\end{figure*}

As said before, \textit{ExtruOnt} is divided in five modules aiming to describe several characteristics of an extruder machine (see Fig. \ref{fig:ExtruOntDiag}).

In the following, the key features of each module are presented.

\subsection{components4ExtruOnt} \label{c4e}
The \textit{components4ExtruOnt}\footnote{\url{http://bdi.si.ehu.es/bdi/ontologies/ExtruOnt/components4ExtruOnt.owl}} module is the main module of the \textit{ExtruOnt} ontology and is intended to describe the components of an extruder. 
According to \cite{giles2004extrusion}, five major systems can be distinguished in an extruder: 
\begin{itemize}
    \item Drive system.
    \item Feed system.
    \item Screw, barrel and heating system.
    \item Head and die assembly.
    \item Control system.
\end{itemize}

Moreover, the components of each one of these systems are explained. For instance, the drive system is composed of motor, gear box, bull gear, and thrust bearing; and the head and die assembly contains the  head, die/nozzle, breaker plate and filters/screens. This analysis of the components of the extruder was used as base to create the \textit{components4ExtruOnt} module.

\begin{figure*}[h]
    \centering
    \includegraphics[width=1\textwidth]{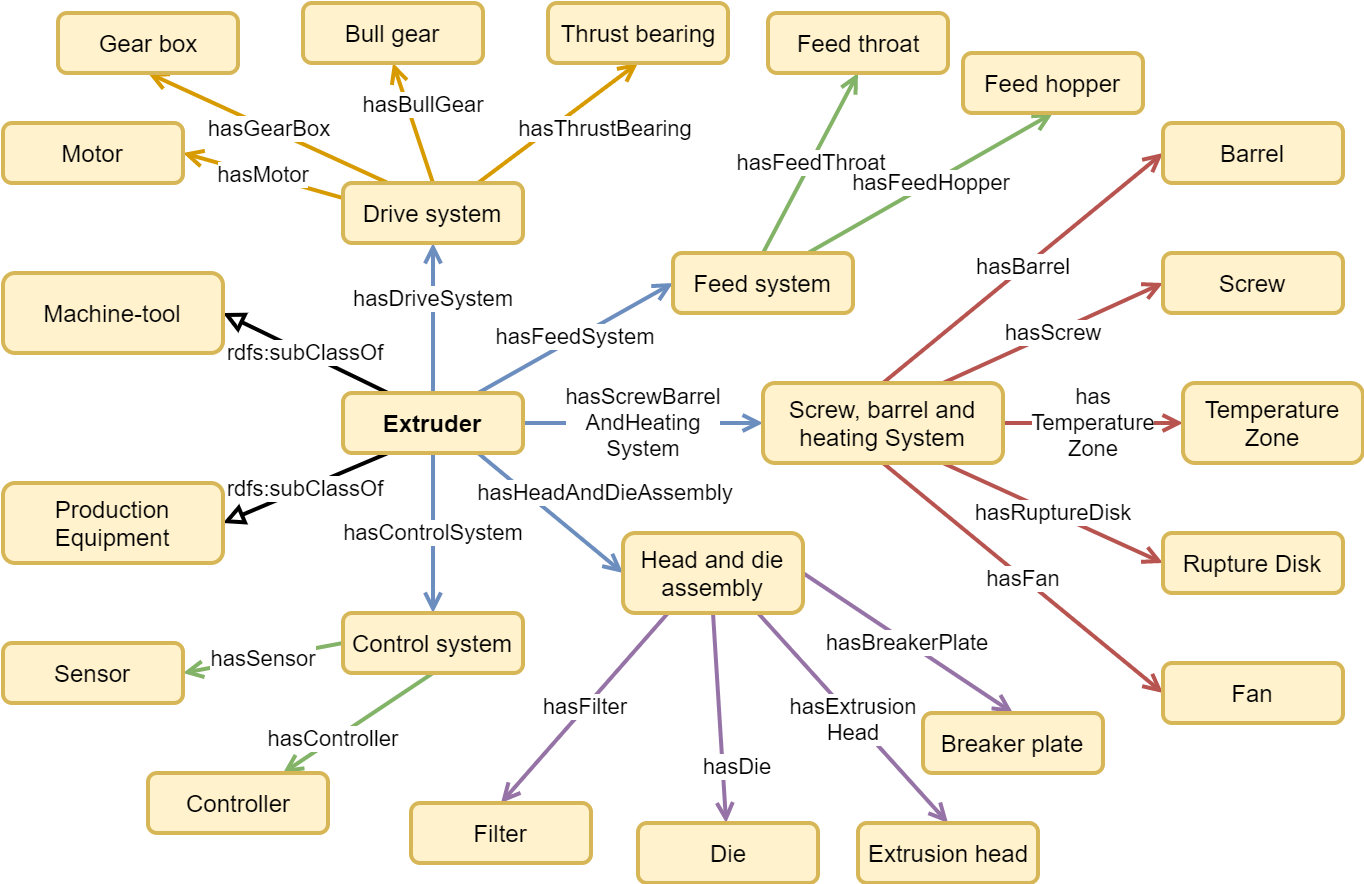}
    \caption{Some components of an extruder.}
    \label{fig:partOf}
\end{figure*}

 A new main class called \verb|Extruder| was created for representing the extrusion machine, while the connections between the extruder and its systems and components were made using the \verb|hasPart| object property of the  \verb|PartOf|\footnote{\url{http://www.ontologydesignpatterns.org/cp/owl/partof.owl}} ontology design pattern. Moreover, custom-made specializations of \verb|hasPart| were created to relate specific components, e.g., \verb|hasBarrel|, \verb|hasScrew| and \verb|hasHeaterBand|. The parthood relations of the extruder and its components are shown in Fig. \ref{fig:partOf}. To facilitate integration with other domain ontologies, the terms \verb|saref4inma:ProductEquip-| \verb|ment|\footnote{\url{https://w3id.org/def/saref4inma}} and 
 \verb|MASON:Machine-tool|\footnote{\url{https://sourceforge.net/projects/mason-onto/}}
 were included as superclasses of \verb|Extruder|.

\begin{figure}[h]
    \centering
    \includegraphics[width=0.3\textwidth]{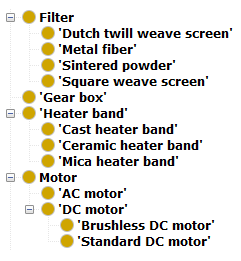}
    \caption{Excerpt of the class hierarchy of the components.}
    \label{fig:SubClass}
\end{figure}

Moreover, the specialization of each component was represented using \verb|rdfs:subClassOf| relations. An example is illustrated in Fig. \ref{fig:SubClass}.

\begin{figure*}[h]
    \centering
    \includegraphics[width=1\textwidth]{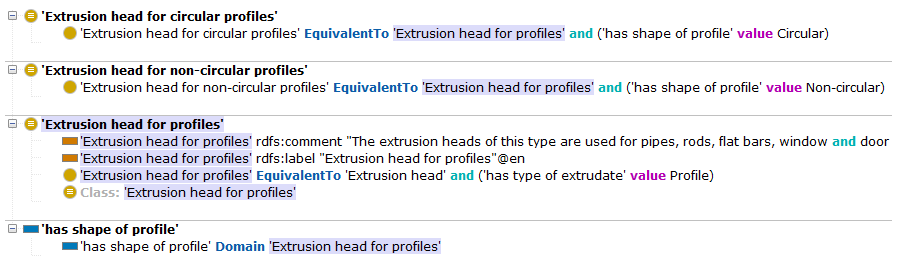}
    \caption{Definition of the Extrusion head for profiles.}
    \label{fig:headUse}
\end{figure*}

\begin{figure}[h]
    \centering
    \includegraphics[width=0.43\textwidth]{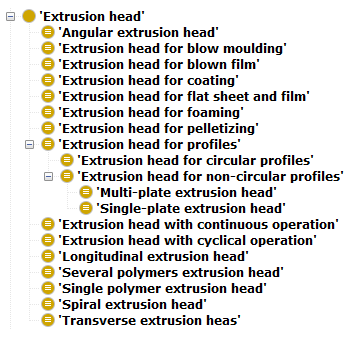}
    \caption{Subclasses of Extrusion head.}
    \label{fig:headSubClass}
\end{figure}

With respect to the extrusion head, the classification that can be found in \cite{sikora2008design} was used to provide a detailed representation of this component. Figs. \ref{fig:headUse} and \ref{fig:headSubClass} exemplify this representation.

Among others, the following competency questions are resolved with the \textit{components4ExtruOnt} module:
\begin{itemize}
    \item CQ1.1: How many heater bands does the extruder E01 have?
    \item CQ1.2: What kind of extrusion head does the extruder E02 have?
    \item CQ1.3: Is the machine E03 a single or double screw extruder?
    \item CQ1.4: Is the extruder E04 powered by an AC motor?
    \item CQ1.5: Is this extruder E05 suitable to process plastic pellets?
    \item CQ1.6: Can the extruder E06 process multiple polymers?
\end{itemize}

A SPARQL query to answer the competency question CQ1.4 is as follows\footnote{We assume that the query is executed after inferences are provided by a reasoner (This applies for all the examples in this paper.)}:

\begin{Verbatim}[fontsize=\scriptsize]
PREFIX : <http://bdi.si.ehu.es/bdi/ontologies/
    ExtruOnt/Extruder01#>
PREFIX rdf: <http://www.w3.org/1999/02/
    22-rdf-syntax-ns#>
PREFIX c4e: <http://bdi.si.ehu.es/bdi/ontologies/
    ExtruOnt/components4ExtruOnt#>
PREFIX p: <http://www.ontologydesignpatterns.org/
    cp/owl/partof.owl#>
ASK { :E04 p:hasPart ?motor01.
    ?motor01 a c4e:AC_motor
	 }
\end{Verbatim}

As a result, the description of the extruder in the \textit{components4ExtruOnt} module will help novice workers to recognize its different sections and components. Moreover, it will help domain experts to formulate queries, according to their needs, related to the amount of components and their types.

\subsection{spatial4ExtruOnt} \label{s4e}
\begin{figure}[h]
    \centering
    \includegraphics[width=0.45\textwidth]{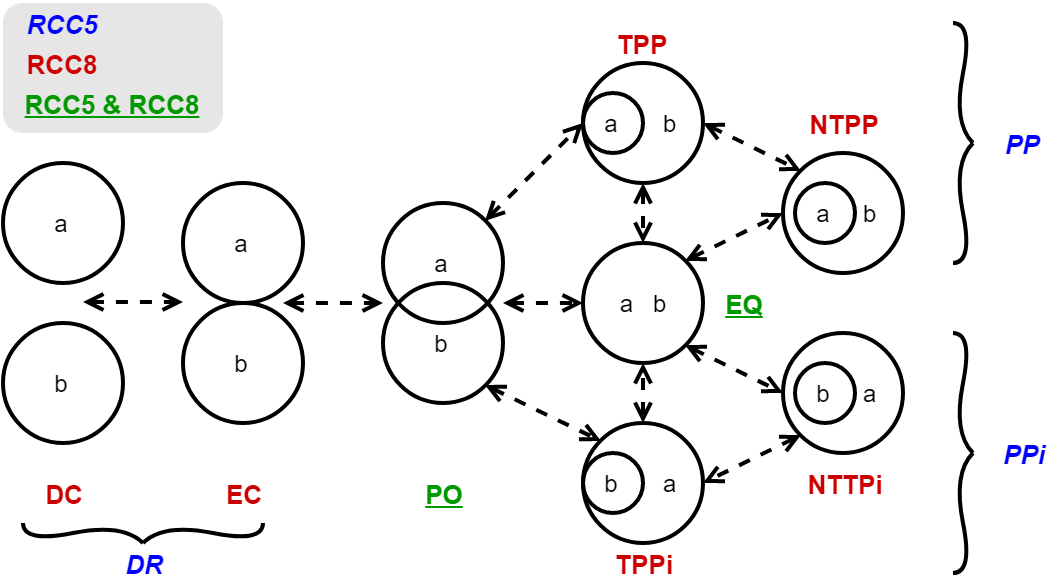}
    \caption{RCC5 and RCC8 relations.}
    \label{fig:rcc8}
\end{figure}

The main representation of RCC is RCC8, which consists of 8 basic relations that are possible between two regions: Equal (EQ), Disconnected (DC), Externally Connected (EC), Partially Overlapping (PO), Tangential Proper Part (TPP), Non-Tangential Proper Part (NTPP), Tangential Proper Part inverse (TPPi) and Non-Tangential Proper Part inverse (NTTPi).  A stripped down version of RCC8 is RCC5, which consists of 5 relations: Equal (EQ), Discrete (DR), Partially Overlapping (PO), Proper Part (PP) and Proper Part inverse (PPi). The graphical representation of RCC5 and RCC8 relations with their mappings are shown in Fig. \ref{fig:rcc8}.

\begin{figure}[h]
    \centering
    \includegraphics[width=0.45\textwidth]{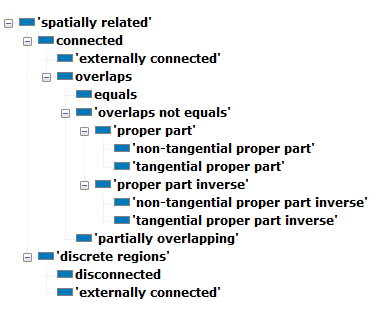}
    \caption{Object property hierarchy in \textit{spatial4ExtruOnt}.}
    \label{fig:rcc8oph}
\end{figure}

For the \textit{spatial4ExtruOnt}\footnote{\url{http://bdi.si.ehu.es/bdi/ontologies/ExtruOnt/spatial4ExtruOnt.owl}} module, a submodule of the GeoSPARQL ontology was used, which contains the \verb|SpatialObject| main class and the object properties referencing to the RCC8 relations. To encourage semantic interoperability, the term \verb|PhysicalObject| 
from DUL ontology\footnote{\url{http://ontologydesignpatterns.org/wiki/Ontology:DOLCE+DnS\_Ultralite}} was included as a superclass of \verb|SpatialObject|. Moreover, a hierarchical object property representation was made including RCC8 relations connected to RCC5 ones, and some more general custom-defined properties. For example, \verb|rcc8tpp| (tangential proper part) is a subproperty of \verb|rcc5pp| (proper part) and, in the same way, \verb|rcc5pp| is a subproperty of the custom-made \verb|overlapsNotEquals| object property. Another example is the following: when two objects overlap, three possible situations can occur: 1) \verb|A| is equal to \verb|B|, 2) \verb|A| partially overlaps \verb|B| and 3) \verb|A| overlaps but is not equal to \verb|B|. This is represented with the \verb|overlaps| object property and three subproperties: \verb|rcc8eq| (equals), \verb|rcc8po| (partially overlapping) and \verb|overlapsNotEquals| (overlaps but not equal). This hierarchy allows a fine-grained classification of spatial relations and can provide detailed results to general questions, e.g., the answer to the question about the objects that overlaps object X will return those objects that are equals, partially overlapping and proper part of object X. The object property hierarchy is shown in Fig. \ref{fig:rcc8oph}.

\begin{figure*}[h]
    \centering
    \includegraphics[width=0.83\textwidth]{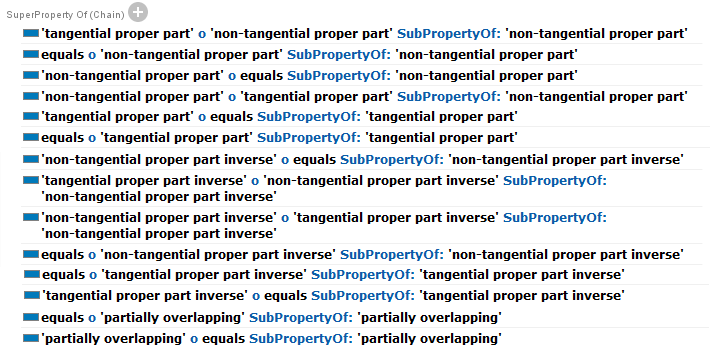}
    \caption{Property chains defined in \textit{spatial4ExtruOnt}}
    \label{fig:propertyChains}
\end{figure*}

RCC8 also defines a composition table where the possible relations between an object A and an object C are indicated based on the relation between object A and B, and the relation between object B and C. However, the OWL 2 DL expressivity level is not sufficient to represent the full table, and for that reason, in \textit{spatial4ExtruOnt} only compositions that yield a single result for the type of relation between objects A and C have been defined in the ontology, more precisely by means of property chains (see Fig. \ref{fig:propertyChains} ).

\begin{figure}[h]
    \centering
    \includegraphics[width=0.4\textwidth]{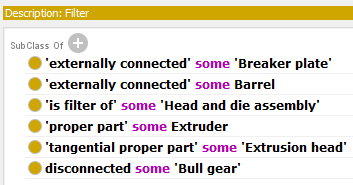}
    \caption{Excerpt of the Filter class description.}
    \label{fig:filter}
\end{figure}

Once the \textit{spatial4ExtruOnt} module was added to \textit{ExtruOnt}, it was possible to describe the spatial connections between the components of the extruder. The classes that describe single components were declared as subclasses of the \verb|SpatialObject| class and the relations between components were made. For example: the filter is externally connected to the barrel and the breaker plate, and it is a tangential proper part of the extrusion head (Fig. \ref{fig:filter}).

With the \textit{spatial4ExtruOnt} module, it is possible to answer several competency questions. These are some of them:
\begin{itemize}
    \item CQ2.1: With which components are the filters FIL01 connected?
    \item CQ2.2: Which components overlap the barrel BAR01?
    \item CQ2.3: Which components are disconnected with the motor M01?
    \item CQ2.4: Which components are monitored in the drive system DS01?
    \item CQ2.5: How many sensors does the barrel BAR02 have?
\end{itemize}

The CQ2.2 competency question is resolved with the following SPARQL query:

\begin{Verbatim}[fontsize=\scriptsize]
PREFIX : <http://bdi.si.ehu.es/bdi/ontologies/
    ExtruOnt/Extruder01#>
PREFIX s4e: <http://bdi.si.ehu.es/bdi/ontologies/
    ExtruOnt/spatial4ExtruOnt#>
SELECT DISTINCT ?component
WHERE { 
  {?component s4e:overlaps :BAR01}
    UNION
  {:BAR01 s4e:overlaps ?component}
}
\end{Verbatim}

The \textit{spatial4ExtruOnt} module will allow novice workers to understand the spatial connections between the different components of an extruder. Furthermore, it will help product designers and domain experts to define the distribution of the components, e.g., the position of the sensors in the head and die assembly. 

\subsection{OM4ExtruOnt} \label{o4e}
The objective of the \textit{OM4ExtruOnt}\footnote{\url{http://bdi.si.ehu.es/bdi/ontologies/ExtruOnt/OM4ExtruOnt.owl}} module is to provide the terms that are necessary to describe the features of the components. This is an important step in the representation of the extruder, as single components could have different characteristics: a barrel could have different dimensions and manufacturing materials. 

A submodule of the OM ontology was used to create \textit{OM4ExtruOnt}, where only the concepts useful for characterizing the components of the extruder and process were taken into account. As stated before, due to the fact that OM is an ontology in the context of food research, it is common to find concepts like \verb|NumberColor1| and \verb|NumberRottenFlowers| to refer to the avocado color and flower status respectively. Consequently, these concepts were removed keeping only concepts like temperature, speed, size, etc.

\begin{figure}[h]
    \centering
    \includegraphics[width=0.4\textwidth]{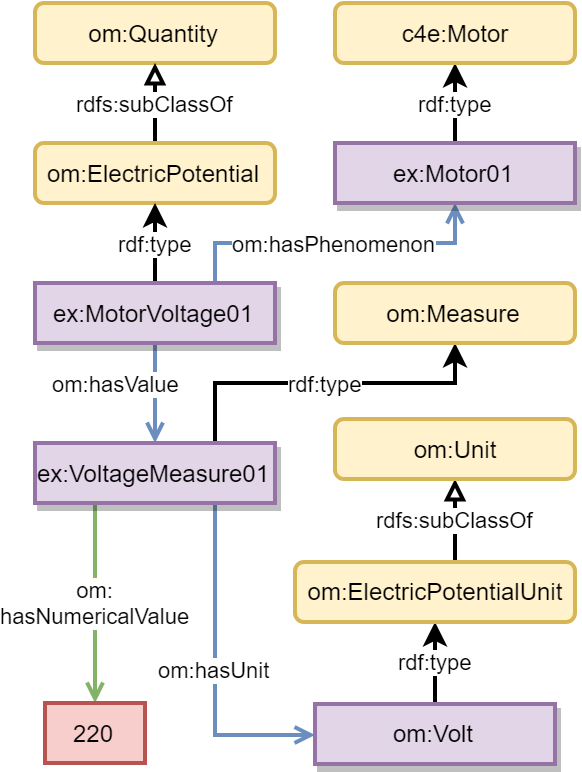}
    \caption{Example of definition of a measure for the feature Motor voltage.}
    \label{fig:motorVoltage}
\end{figure}

The elements of the \textit{OM4ExtruOnt} module can be connected to the elements of the \textit{components4ExtruOnt} module by means of the object property \verb|hasPhenom-| \verb|enon|, which links a measure made for a feature with the object to which the measure applies. For example, in Fig. \ref{fig:motorVoltage} a measure (\verb|ex:VoltageMeasure01|) of the motor voltage (\verb|ex:MotorVoltage01|) of a specific motor (\verb|ex:Motor01|) is represented, which in this case takes the value of 220 volts.

Once the features of the components are defined using the \textit{OM4ExtruOnt} module, it is possible to answer more competency questions, such as:

\begin{itemize}
    \item CQ3.1: What is the diameter of the barrel BAR03?
    \item CQ3.2: What are the optimal operating conditions of the screw SCR01?
    \item CQ3.3: What is the maximum torque produced by the motor M02?
    \item CQ3.4: Does the extruder E07 fit in a space 3 meters wide by 5 meters long?
    \item CQ3.5: What is the bottles-per-hour production rate of the extruder E08?
\end{itemize}

To solve the CQ3.3 competency question a SPARQL query was designed:

\begin{Verbatim}[fontsize=\scriptsize]
PREFIX : <http://bdi.si.ehu.es/bdi/ontologies/
    ExtruOnt/Extruder01#>
PREFIX rdf: <http://www.w3.org/1999/02/
    22-rdf-syntax-ns#>
PREFIX om: <http://www.ontology-of-units-of-
    measure.org/resource/om-2/>
SELECT ?motorTorque01 ?torqueMeasure ?value ?unit
WHERE { ?motorTorque01 a om:Torque.
  ?motorTorque01 om:hasPhenomenon :M02.
  ?motorTorque01 om:hasValue ?torqueMeasure.
  ?torqueMeasure om:hasUnit ?unit;
  om:hasNumericalValue ?value.  
}
\end{Verbatim}

On the one hand, the definition of the features of the components 
made on the \textit{OM4ExtruOnt} module will contribute to the novice workers' awareness of the maximum operating condition of the components. On the other hand, it provides a tool for domain experts to annotate the features of the components, gathered from the design process facilitating the preparation of their specification.

\subsection{3D4ExtruOnt} \label{34e}

\begin{figure}[h]
    \centering
    \includegraphics[width=0.47\textwidth]{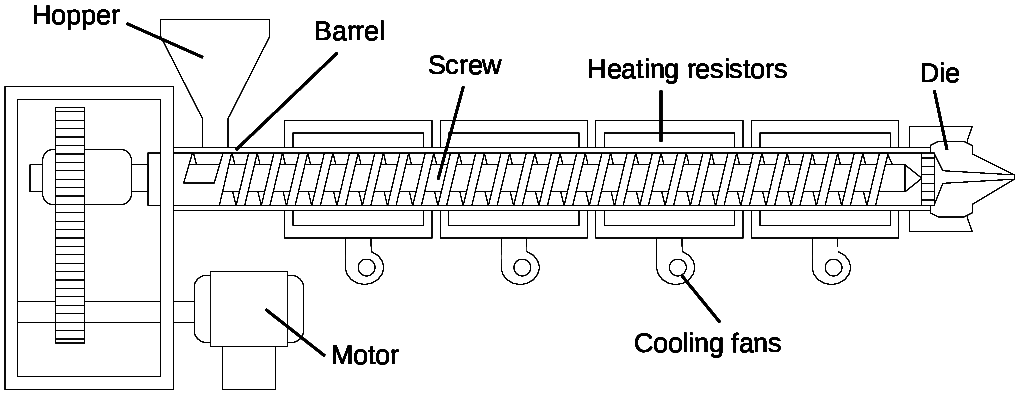}
    \caption{2D representation of the components of an extruder.}
    \label{fig:2DExtruder}
\end{figure}

\begin{figure}[h]
    \centering
    \includegraphics[width=0.45\textwidth]{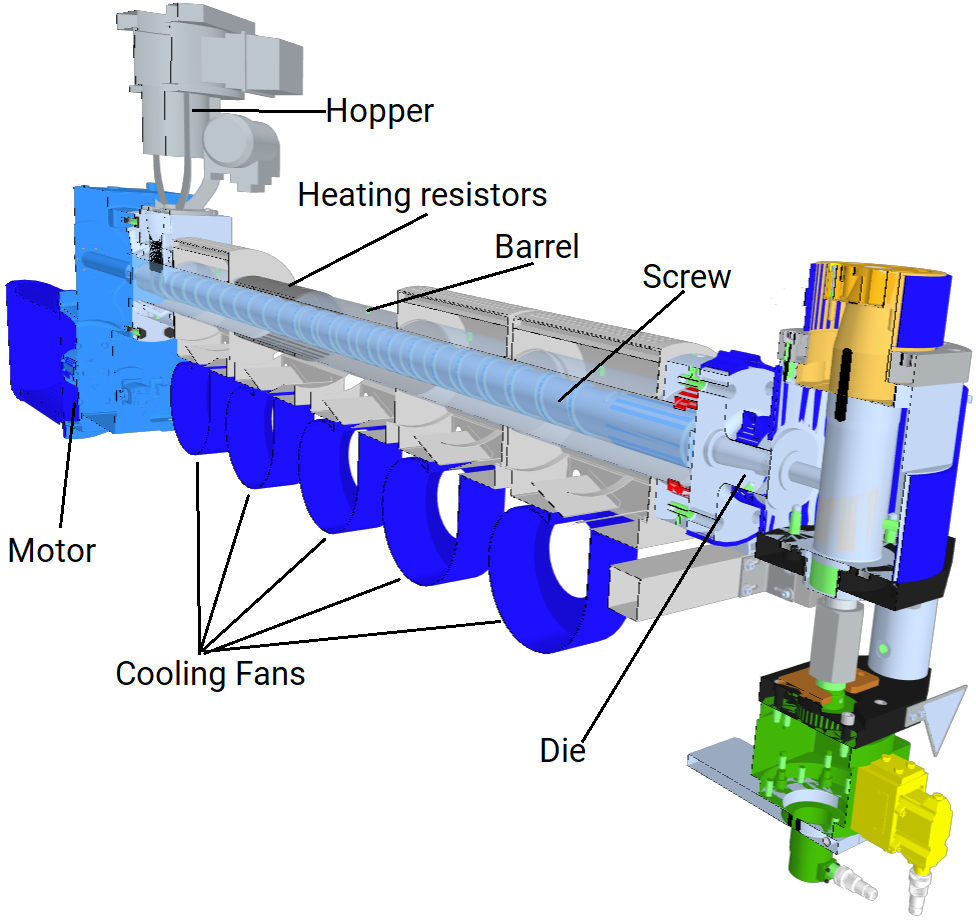}
    \caption{3D representation of the components of an extruder.}
    \label{fig:3DExtruder}
\end{figure}

The graphic representation of an extruder permits to visually understand/observe the positioning of each component that is part of it. Many images of extruders can be found in books, articles, brochures and websites. However, the limitations of a 2D environment makes it difficult to visualize the exact position of the components. Thus, the understanding of an extruder is limited due to the lack of interaction, and the viewer is restricted to the bi-dimensional expressiveness of the author (Fig. \ref{fig:2DExtruder}). On the contrary, a 3D representation of an extruder allows to improve the viewer's interaction, facilitating to move, rotate, zoom in and zoom out. This advantage provides each user with a personalized experience (Fig. \ref{fig:3DExtruder}). 

The purpose of the \textit{3D4ExtruOnt}\footnote{\url{http://bdi.si.ehu.es/bdi/ontologies/ExtruOnt/3D4ExtruOnt.owl}} module is to provide terms for describing the position of each single component in the extruder, in a way that each single component model can be located in a 3D canvas. 

X3D is a royalty-free open standards file format and run-time architecture to represent and communicate 3D scenes and objects, which is approved for the International Standards Organization (ISO). With a set of rich features, X3D can be used in scientific visualization, CAD and architecture, training and simulation, etc. and supports:

\begin{itemize}
    \item 3D graphics and programmable shaders
    \item 2D graphics
    \item CAD data
    \item Animation
    \item User interaction
    \item Navigation

\end{itemize}

\begin{figure}[h]
    \centering
    \includegraphics[width=0.4\textwidth]{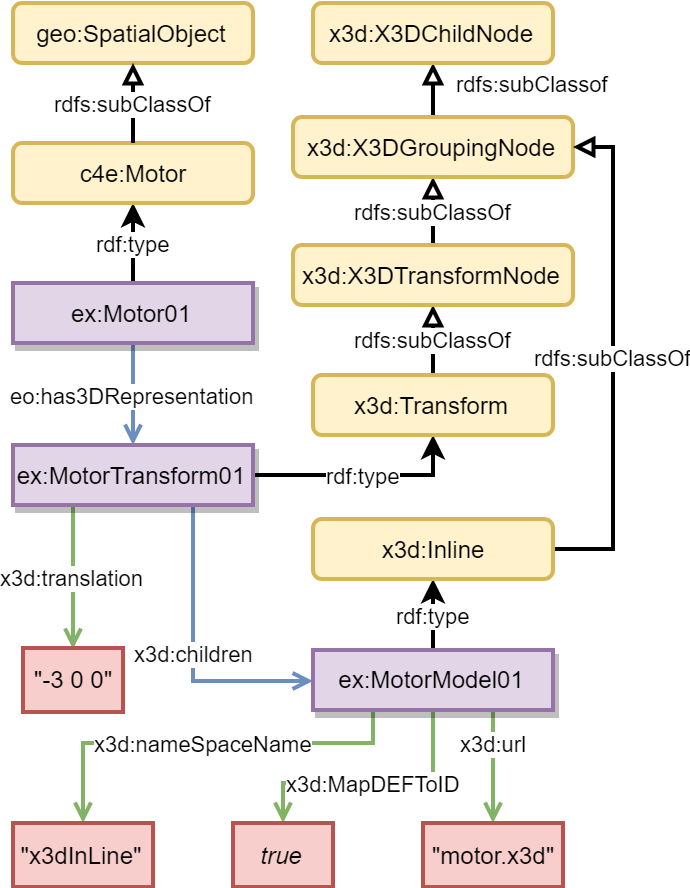}
    \caption{Definition of motor model location in a 3D canvas.}
    \label{fig:3DMotor}
\end{figure}

The selected 3DMO ontology contains a complete X3D definition. To build the \textit{3D4ExtruOnt} module, only the section referring to the 3D object positioning was selected.
To connect the elements of the \textit{3D4ExtruOnt} module with the elements of the \textit{components4ExtruOnt} module, a new \verb|has3DRepresen-| \verb|tation| object property was included, whose range is the X3D \verb|Transform| class and the domain is the \verb|SpatialObject| class, previously mentioned. 
\verb|Transform| class provides the \verb|translation| property where the x, y and z coordinates, referring to the position of a 3D model in a canvas, can be specified. The \verb|Inline| class allows to load different external 3D file formats (obj, stl, collada, fbx, etc.) by using the \verb|url| property to specify the path to the resource location.
An example of the 3D positioning of the motor is shown in Fig. \ref{fig:3DMotor}. 

Now, it is possible to answer competency questions referring to 3D object positioning, for example:

\begin{itemize}
    \item CQ4.1: Which components of extruder E11 can not be located in a 3D canvas?
    \item CQ4.2: What are the modeling and position of the feed hopper FH01?
\end{itemize}

The following SPARQL query can be used to answer the competency question CQ4.2:

\begin{Verbatim}[fontsize=\scriptsize]
PREFIX : <http://bdi.si.ehu.es/bdi/ontologies/
    ExtruOnt/Extruder01#>
PREFIX rdf: <http://www.w3.org/1999/02/
    22-rdf-syntax-ns#>
PREFIX e: <http://bdi.si.ehu.es/bdi/ontologies/
    ExtruOnt/ExtruOnt#>
PREFIX x3d: <http://purl.org/ontology/x3d/>
SELECT ?position ?nameSpace ?id ?url
WHERE { :FH01 e:has3DRepresentation ?hopper3d.
  ?hopper3d a x3d:Transform;
    x3d:translation ?position;
    x3d:children ?model3d.
  ?model3d a x3d:Inline;
    x3d:nameSpaceName ?nameSpace;
    x3d:MapDEFToID ?id;
    x3d:url ?url.
}
\end{Verbatim}

The \textit{3D4ExtruOnt} module will help domain experts in the design process of components, by providing the required information to position 3D models of components in a scene. Moreover, the detection of faults or collisions will be facilitated. Furthermore, it will help novice workers to understand the physical appearance of single components and recognize them in real-world scenarios.\\

\subsection{sensors4ExtruOnt} \label{se4e}
This module is intended to enable domain experts to gain a greater value and insights out of the captured data from the sensors of the extruders, in order to keep trace of the performance of the extruder and allowing to detect possible future faults.

The \textit{sensors4ExtruOnt}\footnote{\url{http://bdi.si.ehu.es/bdi/ontologies/ExtruOnt/sensors4ExtruOnt.owl}} module imports the \textit{SOSA/ SSN} \cite{haller2018sosa} and \textit{OM4ExtruOnt} ontologies. The class \verb|Sensor| was created as a specialization of \verb|sosa:| \verb|Sensor| .  Two properties were added to this class: \verb|indicatorId| (the identifier of the sensor) and \verb|sensorName| (the name of the sensor). Moreover, two main subclasses of Sensor were defined: \verb|Bool-| \verb|eanSensor| and \verb|DoubleValueSensor| to represent sensors that capture true/false data and numerical data respectively.
Finally, these two subclasses were specialized for describing  more specific type of sensors, more precisely sensors for observing: whether a resistor is on or off, whether a fan is on or off, the level and composition of the additive, the number of bottles made in a shift, the feed rate of the polymer, the melting temperature of the polymer, the power consumption of the motor, the pressure in the pressurized zones of the extruder, the speed of the rotational components, the temperature, the thickness of the extrudate and the viscosity of the extrudate.

\begin{figure*}[h]
    \centering
    \includegraphics[width=0.9\textwidth]{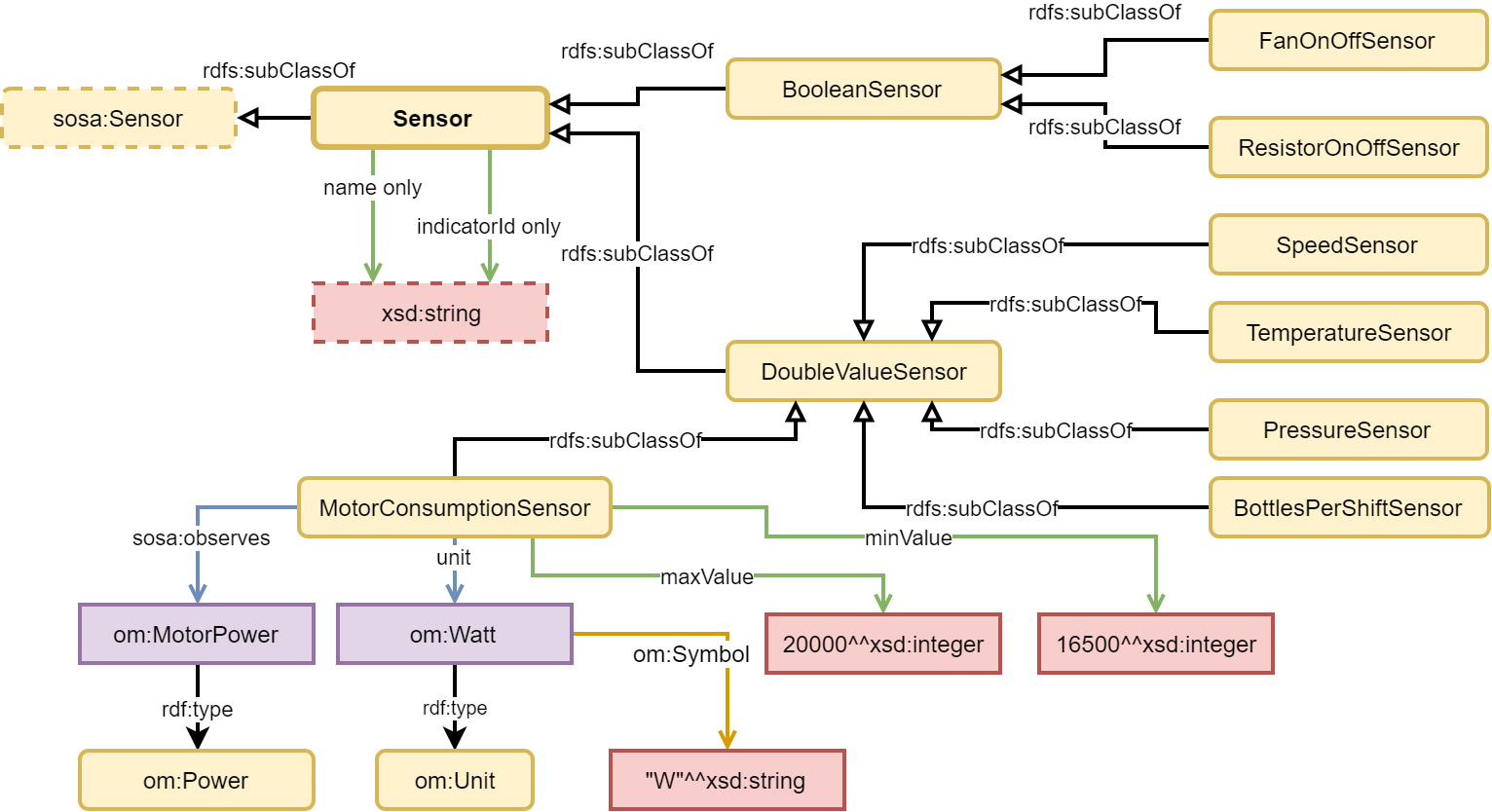}
    \caption{Excerpt of the \textit{sensors4ExtruOnt} module showing some classes and properties related to sensors.}
    \label{fig:Sensors4ExtruOnt}
\end{figure*}

The observable property for each sensor type is indicated by \verb|sosa:observes|. For example, the observable property of a \verb|MotorConsumptionSensor| is \verb|Power| (imported from \textit{OM4ExtruOnt}) and its unit is \verb|Watt|, an individual of \verb|PowerUnit|.
Each sensor type is related to the type of observation that it makes through the \verb|sosa:madeObservation| property. For each observation, its value and timestamp are indicated by properties \verb|sosa:hasSimpleResult| and \verb|sosa:ResultTime| respectively. 
The annotations made in the data and the descriptions in the module can be used to generate a customized and semantically enriched chart to visualize the data. For example, when a sensor is defined as an individual of \verb|MotorConsumptionSensor| class, it can be inferred that it captures values in Watts, its symbol is W and its optimal operational values are between 15,600 and 20,000 units. This information can be used to select the most convenient visual representation of the data, improving the analysis and user experience.
An excerpt of the module can be found in Fig. \ref{fig:Sensors4ExtruOnt}.

\begin{figure*}
    \centering
    \includegraphics[width=0.9\textwidth]{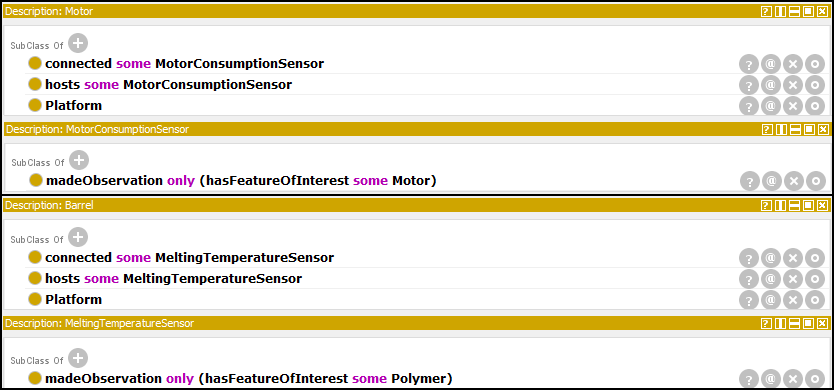}
    \caption{Excerpt of the descriptions of classes Motor, MotorConsumptionSensor, Barrel and MeltingTemperatureSensor}
    \label{fig:sensorsHostsAndHasFeatureOfInterest}
\end{figure*}

In order to indicate the spatial location of a sensor in the extruder the terms described in the module \textit{spatial4ExtruOnt} can be used. In addition, the parts of the extruder (described in the module \textit{components4ExtruOnt}) that host sensors can be seen as \verb|sosa:Platform|s, and linked to them via the object property \verb|sosa:hosts|. Finally, the feature of interest of the observations of each type of sensors has been indicated using the property \verb|sosa:hasFeatureOf-| \verb|Interest|. For example, in the case of a \verb|MotorCon-| \verb|sumptionSensor| the motor of the extruder is both its platform and its feature of interest, while in the case of a \verb|MeltingTemperatureSensor| the platform is the barrel of the extruder and its feature of interest is the polymer used in that extrusion process (see Fig. \ref{fig:sensorsHostsAndHasFeatureOfInterest}).

With the addition of this module, a selection of competency questions can be solved, among others:
\begin{itemize}
    \item CQ5.1: What properties are observed by the sensors located in the extrusion head EH01?
    \item CQ5.2: What is the unit of measurement used by the motor consumption sensor MCS01?
    \item CQ5.3: Where is the melting temperature sensor located in extruder E08?
    \item CQ5.4: What is the identifier of the temperature sensor in extrusion head EH02?
    \item CQ5.5: When was the first and last observation made by sensor SN01?
    \item CQ5.6: What was the average, maximum and minimum value of the observations in a day for the sensor SN02?
    \item CQ5.7: How many observations from torque sensor SN03 are outside the optimal values?
    \item CQ5.8: How long was the maximum period of extruder E09 inactivity during the last week?
    \item CQ5.9: At what times during August 21st, 2018 and August 22nd, 2018 did the melting temperature exceed the maximum optimal operational value in extruder E10?
\end{itemize}

A SPARQL query to answer the CQ5.9 competency question is presented as follows:

\begin{Verbatim}[fontsize=\scriptsize]
prefix :<http://bdi.si.ehu.es/bdi/ontologies/
    ExtruOnt/Extruder01#>
prefix sosa: <http://www.w3.org/ns/sosa/>
prefix xsd: <http://www.w3.org/2001/XMLSchema#>
prefix sn4e: <http://bdi.si.ehu.es/bdi/ontologies/
    ExtruOnt/sensors4ExtruOnt#>
PREFIX p: <http://www.ontologydesignpatterns.org/
    cp/owl/partof.owl#>
select ?resultValue ?resultTime
where {
  :E10 p:hasPart ?barrel01.
  ?barrel a c4e:Barrel .
  ?barrel sosa:hosts ?meltingTempSn01 .
  ?meltingTempSn01 a sn4e:MeltingTemperatureSensor;
    sosa:madeObservation ?obs;
    sn4e:maxValue ?maxValue.
  ?obs sosa:hasSimpleResult ?resultValue ;
    sosa:resultTime ?resultTime .
  filter(?resultValue > ?maxValue) .
  filter((xsd:dateTime(?resultTime) >=
    "2018-08-21T00:00:00.000Z"^^xsd:dateTime) &&
    (xsd:dateTime(?resultTime) <=
    "2018-08-22T23:59:59.999Z"^^xsd:dateTime))
}
order by asc(?resultTime)
\end{Verbatim}

The \textit{sensors4ExtruOnt} module allows domain experts to analyze and keep trace of sensors data in a structured way, retaining important relations and properties between the data, sensors and components of an extruder, which can be valuable in a future failure prediction process.

\section{Evaluation} \label{eva}
Once the \textit{ExtruOnt} ontology was developed, in order to check its quality, two evaluation goals were considered: \textit{Domain coverage} to see in which extent it covered the considered extrusion domain, and   \textit{Quality of the modeling} in terms of the design and development process and in
terms of the final result. 
The third goal identified by NeOn (\textit{Suitability for an application/task}) will be considered once software artifacts whose core element is the \textit{ExtruOnt} ontology (see section \ref{conclusion}) are built. Moreover, the passing of time will allow to evaluate the ontology regarding the goal of \textit{Adoption and use}.
During the evaluation process, the ontology  was also assessed by three types of persons: 1) A R\&D director of a company that develops machines that produce bottles based on an extrusion process, who we work closely with. This person also provides us real data captured from the machines developed 
by his company. 2) A director of an IBDS (Industrial Big Data Services) Provider company. IBDS is an ITS (Information Technology Supplier) company that  supplies  manufacturers with the required technology and services to smartize their manufacturing businesses. Thus, IBDS Providers constitute a fundamental
agent in industrial scenarios where there is an interest in adopting Smart Manufacturing approaches. 3) An expert in developing and managing ontologies who works in a technology center specialized in the industrial domain.


\subsection{Domain coverage} \label{coverage}

Using the non-ontological resources and reusing some other existing ontologies related to the dimensions considered in the ontology, a first version of \textit{ExtruOnt} ontology was built. Then, after a rigorous discussion process with the three experts, who evaluated the correctness and usefulness of the described information in the ontology, it was redefined and some new terms were incorporated and some others were eliminated. Thus, R\&D director of the company, based on his knowledge about the extrusion process, evaluated the semantic quality of the ontology. For example, he suggested to eliminate the three types of categories that we defined related to type of heads (that appear in the non-ontological resource regarding extrusion heads) and refer them through the definition of new features in the existing extrusion head term (for example, shape of profile and quantity of plates) in order to avoid some ambiguities in the representation.The director of an IBDS, based on his acquired knowledge by providing smart manufacturing services to different types of manufacturing companies, evaluated to what extent the ontology could be adapted and used in other manufacturing scenarios. Considering his comments we saw interesting to deal with two upper ontologies: DUL and MASON (the last one focused on the manufacturing domain), because they contain terms that could be relevant in other scenarios, 
for example \textit{process} and \textit{operation} terms to describe the logistics, schedule and maintenance operations in a factory.
Finally, the expert on ontologies evaluated the quality of the alignments with existing ontologies. In this sense, he suggested the alignments with SAREF4INMA instead of 
SAREF, as was our first approach.
In the final version of the \textit{ExtruOnt} ontology, regarding the main concepts described in 
the non-ontological resources, 125 terms were included, and regarding those related to the extrusion head, 32 were included; covering the 95\% of the vocabulary. The remaining 5\% corresponds to terms out of the ontology scope or without significant value (e.g., parts of obsolete extruder models). 
Evaluation against a gold standard was not possible because after performing a thorough search we could not find a gold standard source to compare. Nevertheless, we will continue with the search process and, as soon as we find it, an additional evaluation step will be performed to reinforce the adaptability and reuse tests made to the ontology.

\subsection{Quality of the modeling} \label{quality}

This evaluation goal focuses on the quality of the ontology and can be assessed using a wide range of approaches. In this section we focus on ontology metrics, in common pitfalls in the ontology development process and 
in the contrast of some defined
criteria used for the evaluation of the ontology during the development process.
We selected these approaches because, using all three, a fairly accurate picture of the ontology quality can be obtained. 

\subsubsection{Ontology metrics} \label{metrics}

The basic ontology metrics, including amount of axioms, classes, properties and individuals in the ontology, were extracted from Prot\'{e}g\'{e}. They are listed in Table \ref{t3}.
A schema and graph metrics comparison with other ontologies of the manufacturing domain is listed in Table \ref{tschema}. The data was extracted using OntoMetrics\footnote{\url{https://ontometrics.informatik.uni-rostock.de/ontologymetrics/index.jsp}}. As it can be seen, the metrics for \textit{ExtruOnt} remain in the range of values of other well-known manufacturing domain ontologies. Some metrics like \textit{Inheritance Richness} and \textit{Equivalence Ratio} present a moderate high value due to the semantic interoperability level achieved, i.e., the amount of reused ontologies. However, comparing specific metrics like \textit{tCardinality}, \textit{Depth} and \textit{xtBreadth} would be unfair since the level of abstraction of the compared ontologies differs. 

\begin{table*}
\caption{Ontology metrics} \label{t3}
\begin{tabular}{lrrrrrH}
\hline
    \textbf{Metrics} & \textbf{Components} & \textbf{Spatial} & \textbf{OM} & \textbf{Sensors} & \textbf{3D} & \textbf{ExtruOnt}  \\
\hline
    Axiom                       & 1010  & 378   & 3740      & 775       & 111   & 6021  \\
    Logical axiom count         & 506   & 88    & 1946      & 199       & 36    & 2779  \\
    Declaration axioms count    & 167   & 40    & 477       & 113       & 25    & 822   \\
    Class count                 & 80    & 1     & 107       & 52        & 8     & 248   \\
    Object property count       & 60    & 15    & 17        & 38        & 1     & 131   \\	
    Data property count         & 0     & 0     & 11        & 9         & 13    & 33    \\	
    Individual count            & 17    & 0     & 308       & 7         & 0     & 332   \\
    Annotation Property count   & 19    & 28    & 39        & 21        & 8     & 115    \\	
    DL expressivity             & SHOIQ & ALRI+ & ALCHON(D) & ALCROIN(D)& ALC(D)& SROIQ(D) \\
\hline
    \textbf{Class axioms} \\
\hline
    SubClassOf                  & 302   & 0 & 148   & 146   & 6 & 602   \\	
    EquivalentClasses           & 25    & 0 & 47    & 0     & 0 & 72    \\	
    DisjointClasses             & 11    & 0 & 0     & 3     & 2 & 16    \\	
    GCI count                   & 0     & 0 & 0     & 0     & 0 & 0     \\	
    Hidden GCI Count            & 1     & 0 & 47    & 0     & 0 & 48    \\
\hline
    \textbf{Object property axioms} \\
\hline
    SubObjectPropertyOf             & 52    & 15    & 1     & 1     & 0 & 69    \\	
    EquivalentObjectProperties      & 0     & 0     & 0     & 0     & 0 & 0     \\
    InverseObjectProperties         & 25    & 3     & 0     & 14    & 0 & 42    \\
    DisjointObjectProperties        & 0     & 0     & 0     & 0     & 0 & 0     \\
    FunctionalObjectProperty        & 0     & 0     & 1     & 2     & 0 & 3     \\
    InverseFunctionalObjectProperty & 0     & 0     & 0     & 1     & 0 & 1     \\
    TransitiveObjectProperty        & 2     & 3     & 0     & 0     & 0 & 5     \\
    SymmetricObjectProperty         & 0     & 9     & 0     & 0     & 0 & 9     \\
    AsymmetricObjectProperty        & 0     & 0     & 0     & 0     & 0 & 0     \\
    ReflexiveObjectProperty         & 0     & 1     & 0     & 0     & 0 & 1     \\
    IrrefexiveObjectProperty        & 0     & 0     & 0     & 0     & 0 & 0     \\
    ObjectPropertyDomain            & 35    & 15    & 15    & 2     & 1 & 68    \\
    ObjectPropertyRange             & 36    & 15    & 16    & 2     & 1 & 70    \\
    SubPropertyChainOf              & 0     & 27    & 0     & 4     & 0 & 31    \\
\hline
    \textbf{Data propery axioms} \\
\hline
    SubDataPropertyOf               & 0 & 0 & 0     & 0     & 0     & 0     \\	
    EquivalentDataProperties        & 0 & 0 & 0     & 0     & 0     & 0     \\
    DisjointDataProperties          & 0 & 0 & 0     & 0     & 0     & 0     \\
    FunctionalDataProperty          & 0 & 0 & 1     & 0     & 0     & 1     \\
    DataPropertyDomain              & 0 & 0 & 11    & 7     & 13    & 31    \\
    DataPropertyRange               & 0 & 0 & 10    & 8     & 13    & 31    \\
\hline
    \textbf{Individual axioms} \\
\hline
    ClassAssertion                  & 21    & 0 & 407   & 7 & 0 & 435   \\
    ObjectPropertyAssertion         & 0     & 0 & 1007  & 0 & 0 & 1007  \\
    DataPropertyAssertion           & 0     & 0 & 282   & 2 & 0 & 284   \\
    NegativeObjectPropertyAssertion & 0     & 0 & 0     & 0 & 0 & 0     \\
    NegativeDataPropertyAssertion   & 0     & 0 & 0     & 0 & 0 & 0     \\
    SameIndividual                  & 0     & 0 & 0     & 0 & 0 & 0     \\
    DifferentIndividuals            & 1     & 0 & 0     & 0 & 0 & 1     \\
\hline
    \textbf{Annotation axioms} \\
\hline
    AnnotationAssertion             & 319   & 229   & 1315  & 410   & 50    & 2323  \\	
    AnnotationPropertyDomain        & 0     & 0     & 0     & 0     & 0     & 0     \\
    AnnotationPropertyRangeOf       & 0     & 0     & 0     & 0     & 0     & 0     \\
\hline
\end{tabular}
\end{table*}

\begin{table*}[htb]
\caption{Schema and Graph metrics comparison} \label{tschema}
\begin{tabular}{llllll}
\hline
\textbf{Schema Metric}&\textbf{ExtruOnt}&\textbf{MaRCO}&\textbf{MASON}&\textbf{MSDL}&\textbf{SAREF4INMA}\\
\hline
    Attribute richness&	0.129921& 0.535484&	0.073171&	0.007418&	0.297297\\
    Inheritance richness&	2.531496& 3.312903&	1.199187&	1.135015&	1.810811\\
    Relationship richness&	0.255787& 0.529115&	0.111446&	0.477816&	0.309278\\
    Attribute class ratio&	0& 0&	0&	0&	0\\
    Equivalence ratio&	0.291339& 0.009677&	0&	0.010386&	0\\
    Axiom/class ratio&	24.192913& 12.43871&	5.926829&	30.317507&	9.081081\\
    Inverse relations ratio&	0.325758& 0.011494&	0.212766&	0.152411&	0.178571\\
    Class/relation ratio&	0.293981& 0.142137&	0.740964&	0.460068&	0.381443\\
\hline
\textbf{Graph Metric}\\
\hline
    Absolute root cardinality&	45& 8&	15&	3&	7\\
    Absolute leaf cardinality&	148& 219&	166&	472&	15\\
    Absolute sibling cardinality&	186& 310&	244&	666&	25\\
    Absolute depth&	478& 1520&	1385&	5766&	58\\
    Average depth&	2.489583& 4.367816&	5.54&	8.479412&	2.230769\\
    Maximal depth&	6& 8&	8&	15&	4\\
    Absolute breadth&	192& 348&	250&	680&	26\\
    Average breadth&	4.682927& 3.702128&	3.164557&	3.4&	2.363636\\
    Maximal breadth&	45& 38&	15&	36&	7\\
    Ratio of leaf fan-outness&	0.582677& 0.706452&	0.674797&	0.700297&	0.405405\\
    Ratio of sibling fan-outness&	0.732283& 1&	0.99187&	0.988131&	0.675676\\
    Tangledness&	0.153543& 0.403226&	0.113821&	0.106825&	0.216216\\
    Total number of paths&	192& 348&	250&	680&	26\\
    Average number of paths&	32.0& 43.5&	31.25&	45.333333&	6.5\\
\hline
\end{tabular}
\end{table*}

\subsubsection{OOPS! evaluation} \label{oops}
The Ontology Pitfall scanner (OOPS!) evaluates an ontology by searching for design pitfalls considered from a catalogue of 41 common pitfalls in the ontology development process, classified in a three level scale: critical, important and minor. Most of them (33 out of 41 pitfalls) can be identified semi-automatically by OOPS!. The initial evaluation of \textit{ExtruOnt} yielded some flaws that were corrected, nonetheless, 2 minor pitfalls remain due to external ontology imports. Table \ref{t2} presents the evaluation summary made by OOPS!.

\begin{table*}[htb]
    \caption{Summary of the OOPS! minor pitfalls for ExtruOnt} \label{t2}
    \begin{tabular}{l p{13cm} }
        \hline
        Code &\textbf{P02: Creating synonyms as classes.} \\
        Description & Several classes whose identifiers are synonyms are created and defined as equivalent (owl:equivalentClass) in the same namespace. \\
        Appears in &\verb|http://www.ontology-of-units-of-measure.org/resource/om-2/CelsiusScale| \\
        &\verb|http://www.ontology-of-units-of-measure.org/resource/om-2/FahrenheitScale| \\
        \hline
        Code &\textbf{P04: Creating unconnected ontology elements.} \\
        Description & Ontology elements (classes, object properties and datatype properties) are created isolated, with no relation to the rest of the ontology. \\
        Appears in &\verb|https://w3id.org/def/saref4inma#ProductEquipment| \\
        &\verb|http://xmlns.com/foaf/0.1/Agent| \\
        &\verb|http://www.owl-ontologies.com/mason.owl#Machine-tool| \\
        &\verb|http://www.w3.org/2006/time#TemporalEntity | \\
        &\verb|http://purl.org/vocommons/voaf#Vocabulary | \\
        \hline
    \end{tabular}
\end{table*}

\subsubsection{Evaluation criteria during the development process} \label{criteria}
Criteria defined in \cite{Vrandecic2009} were used for the evaluation of the ontology during the development process. These criteria are listed below with an explanation of their application in \textit{ExtruOnt}.\\

\begin{itemize}
\item \textbf{Accuracy:} The ontology development process was assisted by three experts. Moreover, the modules of \textit{ExtruOnt} were designed using well supported ontological and non-ontological resources. As evidence, \textit{components4ExtruOnt} was created using two non-ontological resources \cite{giles2004extrusion,sikora2008design}, \textit{spatial4ExtruOnt} is based in the Region Connection Calculus relations, \textit{OM4ExtruOnt} uses a submodule of the well known OM ontology, \textit{3D4ExtruOnt} uses concepts from the 3DMO ontology, which follows an ISO open standard (X3D) and finally, \textit{sensors4ExtruOnt} imports definitions from SOSA/SSN ontology.

\item \textbf{Adaptability:} Each module of \textit{ExtruOnt} can be used individually. Thus, it provides reusability and extensibility, making the ontology easily adaptable to describe other different industrial machines.
For example, to describe a wire drawing machine\footnote{A machine that reduce the diameter of a wire by pulling it through a single or a series of drawing dies.}, a new main ontology should be created (e.g. \textit{WidraOnt}), importing on it four modules from \textit{ExtruOnt}, more precisely, the \textit{spatial4ExtruOnt}, \textit{OM4ExtruOnt}, \textit{sensors4ExtruOnt} and \textit{3D4ExtruOnt} modules, which do not have to be modified since the terms in these modules describe information related to general manufacturing machines. Therefore, only the \textit{components4ExtruOnt} module should be redefined (e.g. \textit{components4WidraOnt}), incorporating to it terms referring to the new components (such as puller, coiling roller, capstan, wire, etc.) that belong to the new machine, importing some terms from \textit{components4ExtruOnt} (such as motor, gearbox, etc.) that are shared between both machines and leaving out some other terms (such as extrusion head, barrel, hopper, etc.) that do not belong to the new machine. The main class \verb|WireDrawingMachine|, which represents the new machine that we want to describe, should be defined as a subclass of \verb|MASON:Machine-tool| and \verb|SAREF4INMA:ProductEquipment| to favour interoperability. The new components should be incorporated under the \verb|owl:Thing| class in the \textit{components4WidraOnt} module and linked to the \textit{spatial4ExtruOnt} module as subclasses of \verb|SpatialObject|. Moreover, the connections between the machine and its components should be made using the \verb|hasPart| object property or new custom-made subproperties of \verb|hasPart|. In this way, it is possible to describe the spatial and parthood relations between components of the new machine, for example:

\begin{Verbatim}[fontsize=\scriptsize]
prefix C4W: <http://bdi.si.ehu.es/bdi/ontologies/
    ExtruOnt/components4WidraOnt#>
prefix C4E: <http://bdi.si.ehu.es/bdi/ontologies/
    ExtruOnt/components4ExtruOnt#>
prefix po: <http://www.ontologydesignpatterns.org/
    cp/owl/partof.owl#>
prefix geo: <http://www.opengis.net/ont/
    geosparql#>

C4W:WireDrawingMachine po:hasPart C4W:Casptan, 
                                C4E:Motor.
C4W:Casptan geo:rcc8ec some C4E:Motor.
\end{Verbatim}

Which means that the wire drawing machine has the capstan and the motor as parts, and the capstan is externally connected to the motor. Finally, some other minor adaptations should be carried out regarding the linking of the new terms to the concepts defined in the other imported modules, as it was explained for \textit{ExtruOnt}.

\item \textbf{Clarity:} The custom terms defined in all modules of \textit{ExtruOnt} contain non-ambiguous names, labels and comments facilitating the human readability and avoiding confusions and difficulty when the creation of individuals is carried out.

\item \textbf{Completeness:} The \textit{ExtruOnt} Ontology can answer all the competency questions specified in the ORSD document, representing correctly the domain for which it was created.

\item \textbf{Efficiency:} Although the submodule extraction process from extensive ontologies such as OM and the utilization of specific terms in the context reduce the size of \textit{ExtruOnt}, the reasoner execution time keeps too long when multiple extruders are described containing several data from sensors. However, the annotation and querying process can be carried out seamless.

\textbf{Conciseness}: The knowledge contained in the modules \textit{components4ExtruOnt} and \textit{spatial4Ex-}\textit{truOnt} was retrieved from sources that are specific to the domains of extrusion and spatial relations respectively, thus avoiding irrelevant information. Moreover, for the remaining modules, submodules from OM, SOSA/SSN and 3DMO were extracted in the \textit{Design} phase so that \textit{ExtruOnt} incorporates only the concepts and descriptions from those ontologies that are relevant for our domain.

\item \textbf{Consistency:} No inconsistencies were found in \textit{ExtruOnt} when reasoning was performed. The reasoner used was Fact++\footnote{\url{http://owl.man.ac.uk/factplusplus/}}.  
\end{itemize}
We did not evaluate the criterion of \textbf{Organizational fitness} because the ontology has not been deployed yet.

\section{Conclusion and future work} \label{conclusion}
The purpose of this paper is to present the \textit{ExtruOnt} ontology, which contains terms to describe a type of manufacturing machine for performing extrusion processes (extruder). It is constituted by five modules: \textit{components4ExtruOnt} for representing the components of an extruder, \textit{spatial4ExtruOnt} for representing spatial relationships among those components, \textit{OM4ExtruOnt} for representing the features of those components, \textit{3D4ExtruOnt} for  representing 3D models of the components, and \textit{sensors4ExtruOnt} for  representing the data captured by sensors. Although the \textit{ExtruOnt} ontology is focused on extruders, it has been defined in such a way that it can be used as a model for describing other types of manufacturing machines by customizing or replacing some of its modules.

The descriptions contained in the \textit{ExtruOnt} ontology will allow different types of users to familiarize themselves with the extrusion process, to interoperate with other manufacturing companies in an easy way, to create customized 3D images of extruder machines and an assisted exploration of data captured by sensors. 

The \textit{ExtruOnt} ontology has been documented and is available online. It has been evaluated according to two evaluation goals: domain coverage and quality of modeling, and has been assessed by humans and software artifacts. 
The evaluation shows that \textit{ExtruOnt} can provide the answers to the competency questions defined, satisfying the proposed requirements and, therefore, proving that its modules are correctly developed.
Furthermore, it is aligned with related ontologies, facilitating interoperability.

Finally, in addition to the necessary task of maintenance, we will mainly focus the future work on the development of two software artifacts whose core element will be the \textit{ExtruOnt} ontology, in order to measure its performance in practical scenarios. The first artifact will be a Visual Query System, that will provide those advantages that we have mentioned through the paper to distinct types of users that work in the considered smart manufacturing scenario. The second artifact will be a recommender system that taking into account, on the one hand, the requirements of clients interested in buying an extruder machine and, on the other hand, the information described in the \textit{ExtruOnt} ontology, will propose the most suitable extruder and the possible customizations that can be incorporated into it.


\begin{acks}
The authors would like to thank Urola Solutions for their help with information about the extrusion process and for providing real data. This research was funded by the Spanish Ministry of Economy and Competitiveness under Grant No.:
FEDER/TIN2016-78011-C4-2R and the Basque Government under Grant No.: IT1330-19. The work of V\'ictor Julio Ram\'irez-Dur\'an is funded by the contract with reference BES-2017-081193.
\end{acks}



\nocite{label} 
\bibliographystyle{ios1}           
\bibliography{bibliography}        

%

\end{document}